\pdfoutput=1
\documentclass[11pt]{article}

\usepackage[preprint]{acl}

\usepackage{times}
\usepackage{latexsym}

\usepackage[T1]{fontenc}
\usepackage[utf8]{inputenc}

\usepackage{microtype}

\usepackage{inconsolata}

\usepackage{graphicx}

\usepackage[most]{tcolorbox}
\usepackage{subfigure}
\usepackage{amsmath}
\usepackage{adjustbox}
\usepackage{booktabs}
\usepackage{multirow}
\usepackage{makecell}
\usepackage{array}
\usepackage{tablefootnote}

\usepackage{algorithm}
\usepackage{algpseudocode}

\usepackage[]{color}
\usepackage{xcolor}

\newcommand\blfootnote[1]{%
  \begingroup
  \renewcommand\thefootnote{}\footnote{#1}%
  \addtocounter{footnote}{-1}%
  \endgroup
}
\definecolor{red}{RGB}{192,0,0}
\definecolor{green}{RGB}{197,224,180}

\title{Understanding When Tree of Thoughts Succeeds: \\ Larger Models Excel in Generation, Not Discrimination}

\author{
 \textbf{Qiqi Chen\textsuperscript{$\ast$ 1}},
 \textbf{Xinpeng Wang\textsuperscript{$\ast$ 1,2}},
 \textbf{Philipp Mondorf\textsuperscript{1,2}},
 \textbf{Michael A. Hedderich\textsuperscript{1,2}},
 \textbf{Barbara Plank\textsuperscript{1,2}}
\vspace{2pt}
\\
 \textsuperscript{1}MaiNLP, Center for Information and Language Processing, LMU Munich, Germany, \\
 \textsuperscript{2}Munich Center for Machine Learning (MCML), Munich, Germany
\vspace{5pt}
\\
 \texttt{
   chen.qiqi@campus.lmu.de, xinpeng@cis.lmu.de
 }
}

\begin{document}
\maketitle
\begin{abstract}
Tree of Thoughts (ToT) is a reasoning strategy for Large Language Models (LLMs) that employs a generator to suggest reasoning steps and a discriminator to decide which steps to implement. ToT demonstrates strong performance on reasoning tasks, often surpassing simple methods such as Input-Output (IO) prompting and Chain-of-Thought (CoT) reasoning. However, ToT does not consistently outperform such simpler methods across all models, leaving large knowledge gaps on the conditions under which ToT is most beneficial. In this paper, we analyze the roles of the generator and discriminator separately to better understand the conditions when ToT is beneficial. We find that the generator plays a more critical role than the discriminator in driving the success of ToT. Scaling the generator leads to notable improvements in ToT performance, even when using a smaller model as the discriminator, whereas scaling the discriminator with a fixed generator yields only marginal gains. Our results show that models across different scales exhibit comparable discrimination capabilities, yet differ significantly in their generative performance for ToT.
\blfootnote{$^\ast$Lead authors.}
\end{abstract}

\section{Introduction}\label{sec:intro}

Since the introduction of CoT \citep{COT}, which has enhanced the reasoning capabilities of LLMs, numerous new prompting-based methods have been proposed to further support LLM-based reasoning \citep{GOT, AOT, IOT, pot, ranaldi2024MultiLangToT, probabilisticToT, Tree_of_Uncertain_Thoughts, wang2023plan, zhouleast, khotdecomposed}. Among these, the ToT method proposed by \citet{Tree_of_Thoughts} extends the CoT approach into a tree search framework, demonstrating its potential to enhance the reasoning performance of state-of-the-art LLMs,
\begin{figure}[h!]
    \centering
    \includegraphics[width=\linewidth]{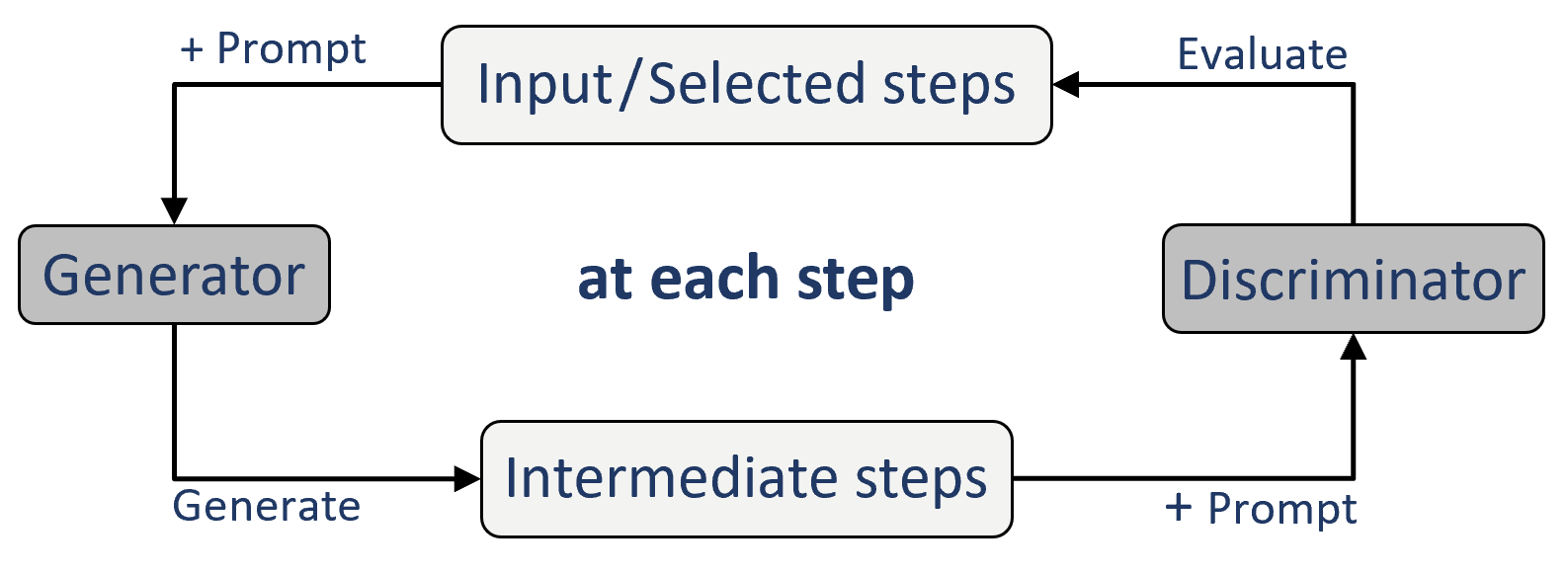}
    \caption{Core Mechanism of ToT. It employs a generator to suggest intermediate steps and a discriminator to decide which steps to take.}
    \label{fig:Core Mechanism of ToT}
\end{figure}
such as GPT-4 \citep{gpt4}, across complex reasoning tasks, including the Game of 24, Creative Writing, and Mini Crosswords \cite{Tree_of_Thoughts}.

ToT is a prompting framework that encourages the model to generate and self-evaluate intermediate reasoning steps.
Theoretically, ToT offers advantages over simpler methods like IO prompting and CoT reasoning through extensive exploration (via the generator) and optimal selection mechanisms (via the discriminator), as shown in Figure \ref{fig:Core Mechanism of ToT}.
However, when ToT is applied to a wider range of LLMs and task types, it has been found to adversely affect the inherent reasoning abilities of weaker LLMs \citep{gtbench}. This reveals that the practical implications of ToT across different model scales remain under-explored. Additionally, \citet{discriminator/MCTS} highlight that advanced planning methods like tree search require high-quality discriminators (accuracy $\geq$ 90\%) to outperform simple re-ranking methods. However, the current discriminative capabilities of most LLMs fall short of this threshold, limiting the effectiveness of such advanced techniques. Their work highlights the importance of the discriminator's capability when using tree-search-based methods; however, their conclusions are drawn based on the evaluation of only one single generator (i.e., CodeLlama-13B-Inst \citep{codellama-13b}).

This study aims to gain deeper insights into whether specific scales of LLMs can benefit from ToT when addressing problems of mathematical and logical reasoning. We systematically compare the performance of ToT against baseline methods, including IO prompting and CoT reasoning, to evaluate ToT's performance across various model scales on challenging mathematical reasoning tasks and natural language-based logical reasoning problems, aiming to identify the conditions under which ToT offers significant improvements. 
Our investigation is framed around three research questions: (1) Does scaling the size of the \textbf{generator} improve ToT's performance? (2) Does scaling the size of the \textbf{discriminator} improve the performance of ToT? (3) Under which conditions does ToT outperform IO and CoT?

Our key findings are:
\begin{itemize}
    \item ToT tends to benefit larger models more than smaller ones.
    \item Scaling the size of the generator results in significant improvements while scaling the size of the discriminator provides only marginal gains.
    \item While models of different scales exhibit similar discriminative abilities when applying ToT, their generation performance varies significantly.
\end{itemize}
By uncovering effective conditions for the successful application of ToT, we aim to aid users in making informed decisions about when to employ the ToT framework and when to use simpler methods to avoid unnecessary resource costs.

\section{Background: Tree of Thoughts}

ToT, as illustrated in Figure \ref{fig:Core Mechanism of ToT} and proposed by \citet{Tree_of_Thoughts}, is a search-based approach designed to enhance reasoning in LLMs by extending the stepwise reasoning of the CoT method \citep{COT} into a tree search paradigm. 
The core mechanism of ToT involves two actors: a generator that proposes multiple intermediate reasoning steps, which are then evaluated by a discriminator to select the most promising ones for subsequent steps in the search process.

ToT systematically assesses candidate steps using search algorithms such as Breadth-First Search \citep{moore1959shortest} and Depth-First Search \citep{tarjan1972depth} to find the optimal path to the best solution. In this context, several terms are crucial for understanding the ToT framework. \textbf{Generation} refers to the process by which a \textbf{generator} produces intermediate steps or explanations, also known as exploration or expansion of the search tree. 
\citet{Tree_of_Thoughts} introduces two key methods for evaluating these steps. The \textbf{Value} method quantifies each step independently, converting it into scalar values (e.g., 1-10) or categorizations (e.g., confident/likely/impossible), based on various evaluative criteria like forward-looking simulations or common sense. The \textbf{Vote} method compares all generated steps through a voting process, selecting the most promising partial solutions when direct assessment of reliability is difficult.

\begin{figure}[htpb]
    \centering
    \includegraphics[width=1\linewidth]{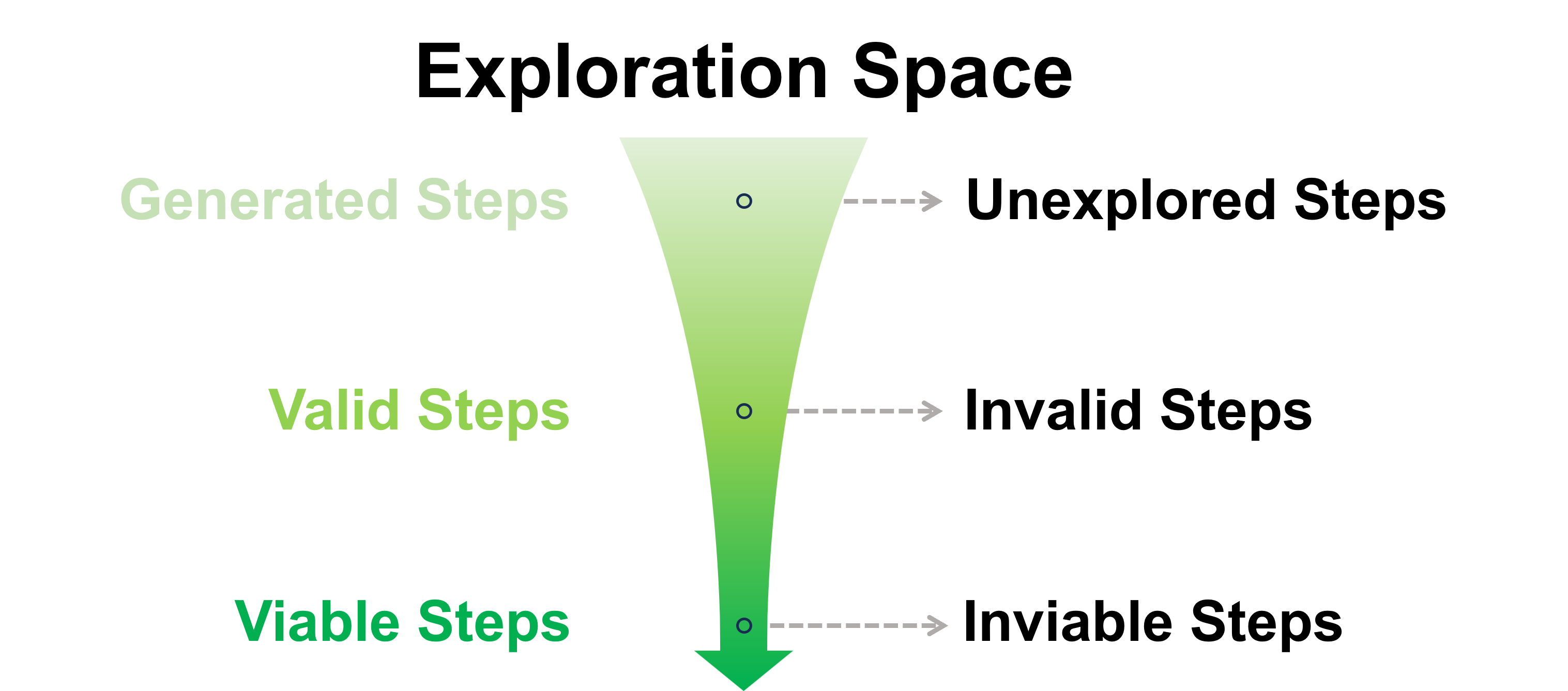}
    \caption{Diagram of relationships among different types of steps in ToT.}
    \label{fig:thoughts}
\end{figure}

Steps generated by the LLM can be categorized as valid or invalid. \textbf{Valid step} adhere to game rules, while \textbf{invalid step} violate them. Among valid steps, \textbf{viable steps} (or promising steps) are those that potentially lead to a solution. \textbf{Inviable steps} (or unpromising steps) are valid but unlikely to lead to success. These relationships are depicted in Figure \ref{fig:thoughts}.

The \textbf{search space} or exploration space refers to the number of choices available at each step. The LLM responsible for evaluating candidate steps is referred to as the \textbf{discriminator} or evaluator. Together, these components form the foundation of ToT's reasoning and search mechanisms.

\section{Experimental Setup}
This section outlines the specific experimental setup, including the tasks and metric used for evaluation, selected LLMs, and the details of ToT implementation for each task. It also covers the baselines used for comparison and the specific prompts employed in the experiments.

\begin{figure}[htbp]
    \centering
    \subfigure[Game of 24]{
        \includegraphics[width=\linewidth]{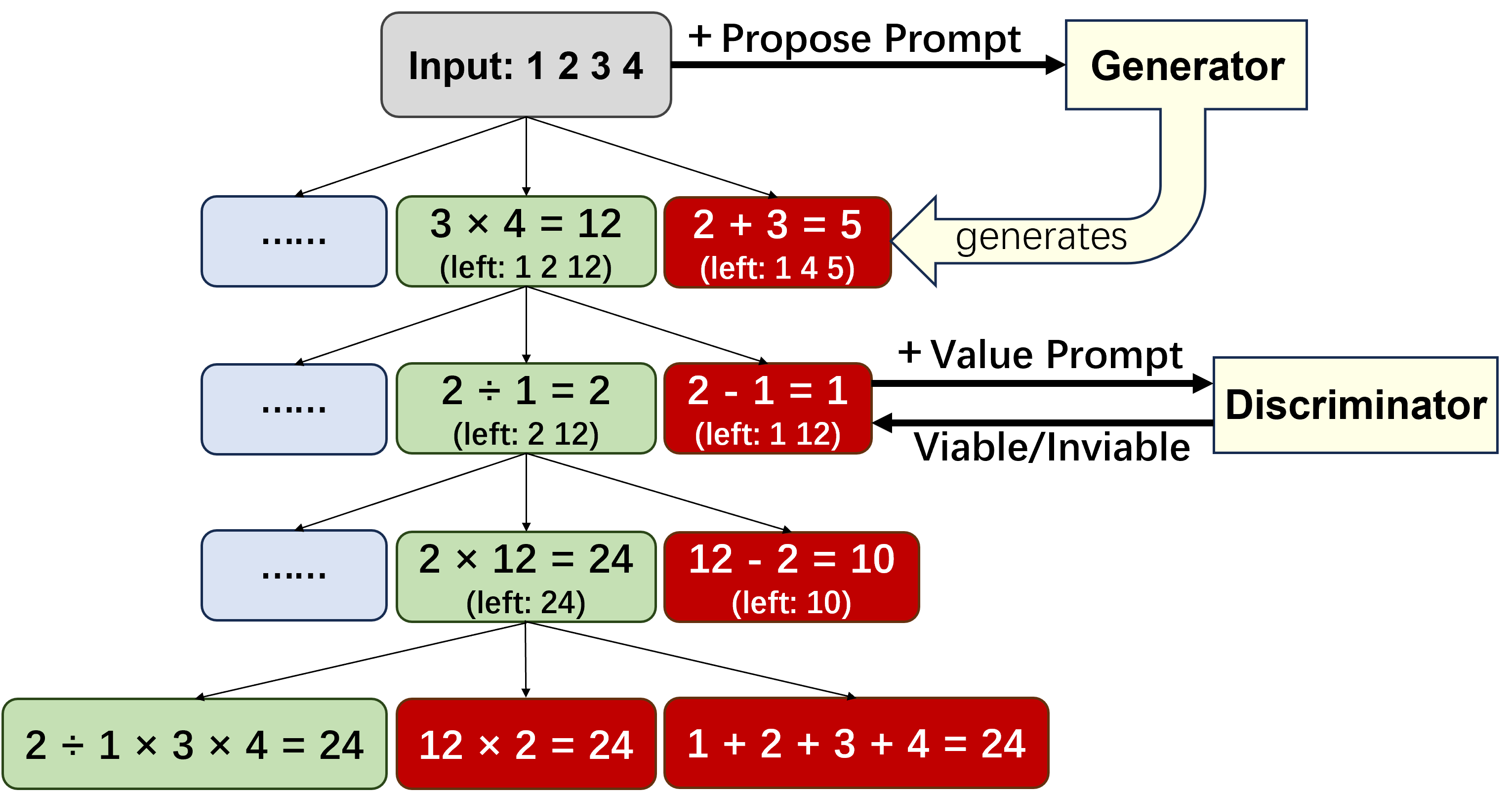}
    }
    \hfill
    \subfigure[Knights and Knaves]{
        \includegraphics[width=0.8\linewidth]{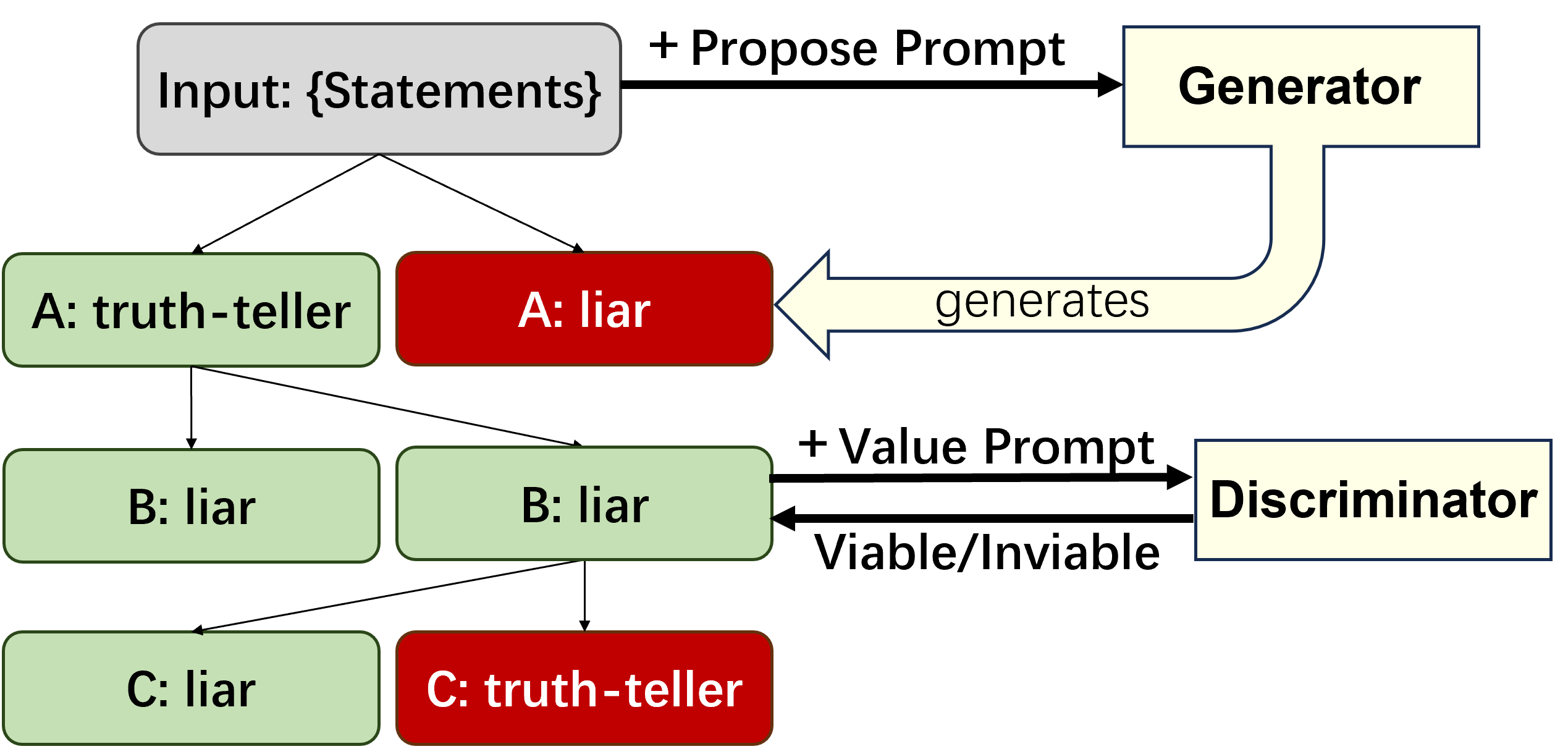}
    }
    \caption{Illustration of Game of 24 and Knights and Knaves under ToT setting. The generator proposes possible intermediate steps, which will be evaluated by the discriminator. We denote the viable/inviable intermediate partial solutions in \textcolor{green}{green}/\textcolor{red}{red}. }
    \label{fig:tasks}
\end{figure}
\subsection{Tasks}\label{sec:tasks}

This study selects two challenging reasoning tasks — Game of 24 and Knights and Knaves — for investigating the efficacy of the ToT approach in mathematical and natural language reasoning tasks. Game of 24 is a mathematical logic puzzle characterized by high decision complexity (refer to Section \ref{sec:g24 rule} for further details). According to the results reported by \citet{Tree_of_Thoughts}, even state-of-the-art LLMs such as GPT-4 \citep{gpt4} using baseline methods achieve a success rate of less than 10\%. In contrast, Knights and Knaves is a natural language-based logic reasoning task that involves extensive hypothesis generation, inference, and backtracking. Despite having a decision complexity of only \(2^{\text{\#characters}}\), \citet{mondorf2024liarliarlogicalmire}'s findings indicate that LLMs employing baseline methods achieved an accuracy of no more than 30\%. Therefore, there is substantial room for improvement in the reasoning capabilities of models for both tasks.

\subsubsection{Game of 24}\label{sec:g24 rule}
\paragraph{Rule}
The Game of 24 is a mathematical reasoning challenge that presents an arithmetic task. The objective is to manipulate exactly four integers within the range of 1 to 13, using basic arithmetic operations, (i.e., $+, -, \times, \div$), to achieve a final result of 24. For instance, given the numbers 1, 2, 3, and 4, a possible solution could be formulated as $(1 + 3) \times (2 + 4) = 24$. As Illustrated in Figure \ref{fig:tasks} (a), in each round of the game, the LLM receives an input of four integers and is expected to provide the correct solution corresponding to this input. This task may possess multiple valid solutions.

Given four numbers, they can be permuted in various sequences. For each permutation, there are three interstitial positions where an operator can be inserted, with four possible operators for each position. Multiplying all the combination possibilities together, we can get a decision complexity of $ 4! \times 4^3 = 1,536$.

\paragraph{Dataset}\label{sec:g24Dataset}
In order to facilitate a more effective comparison between our experimental results and those of \citet{Tree_of_Thoughts}, we utilize the same dataset proposed in their work. This dataset comprises 1,362 games, which are arranged in ascending order of difficulty based on the time taken by humans to solve them. However, this dataset does not include the solutions to the problems. Therefore, we generate all feasible solutions for each task using an algorithm for our experiments (see Appendix \ref{sec:algo}).\footnote{The algorithm and code are included in this \href{https://github.com/mainlp/tot-eval}{GitHub repo}.}

\subsubsection{Knights and Knaves}
\paragraph{Rule}
Knights and Knaves puzzles are a class of logical puzzles in which each character is either a "Knight" or a "Knave". The fundamental rule of these puzzles is that a Knight always tells the truth, meaning that any statement made by a Knight is logically consistent with the facts. In contrast, a Knave always lies, meaning that every statement made by a Knave is false. The objective of a LLM agent is to logically deduce the identity of each character based on their statements (Figure \ref{fig:tasks} (b)).

\paragraph{Dataset}
We utilize the dataset published by \citet{mondorf2024liarliarlogicalmire}, which comprises a total of 2,400 distinct tasks. These tasks are divided into four subsets based on varying numbers of characters (indicating difficulty levels), with \(n = 3, 4, 5, 6\). Each subset contains 600 problems, and every task in the dataset has a unique solution. We use the subset with 3 characters in our experiments.

\subsection{Evaluation Metrics}
\label{sec:metric}
Following the work of \citet{Tree_of_Thoughts}, we use \textit{average success rate} as metric to assess the results of the Game of 24 and Knights and Knaves. 

\paragraph{Average Success Rate}
According to \citet{Tree_of_Thoughts}'s work, the average success rate can be defined as follows: 
\begin{equation}
    \label{eq:asr}
    \text{Average Success Rate} = \frac{\sum_{\text{tasks}} \frac{\text{\# correct answers}}{\text{\# outputs}}}{\text{\# tasks}}
\end{equation}
The average success rate is in the range [0, 1], with 1 indicating that all tasks are successful and 0 indicating that all tasks fail.

Note that the average success rate eliminates the advantage of multiple attempts for the same problem in IO, CoT, and ToT methods, allowing for direct comparison of results across different approaches. This also facilitates comparison with accuracy metrics reported in other studies, even when the number of attempts per task differs.

\subsection{Language Models}
Based on the classification of language model scales in \citet{aqabench}, six open-source language models at three different scales are used in the experiments of this study, which are: 
\begin{itemize}
    \item smaller models (with $\textless$ 10B parameters): gemma-2b-it \citep{gemmateam2024gemma}, Llama-3-8B-Inst \citep{llama3modelcard} and Llama-3.1-8B-Inst \citep{llama3.1},
    \item medium models (with $\geq$ 10B and $\textless$ 50B parameters): gemma-2-27b-it \citep{team2024gemma},
    \item larger models (with $\geq$ 50B parameters): Llama-3-70B-Inst \citep{llama3modelcard} and Llama-3.1-70B-Inst \citep{llama3.1}.
\end{itemize}
In the experiments for \textbf{RQ1} and \textbf{RQ2}, we use Llama-3.1-8B-Inst, gemma-2-27b-it and Llama-3.1-70B-Inst.
In the experiments for \textbf{RQ3}, all models except gemma-2-27b-it are included.
All models used are instruction-tuned models. We query the API provided by Hugging Face\footnote{\url{https://huggingface.co/models}} for model inference. We employ \textit{temperature sampling} as the models' generation strategy in all runs. Unless otherwise specified, the decoding temperature for the model in all experiments is set to 0.7.

\subsection{Baselines \& ToT Oracle}

\begin{figure}[htbp]
    \centering
    \subfigure[IO]{
        \includegraphics[width=0.20\linewidth]{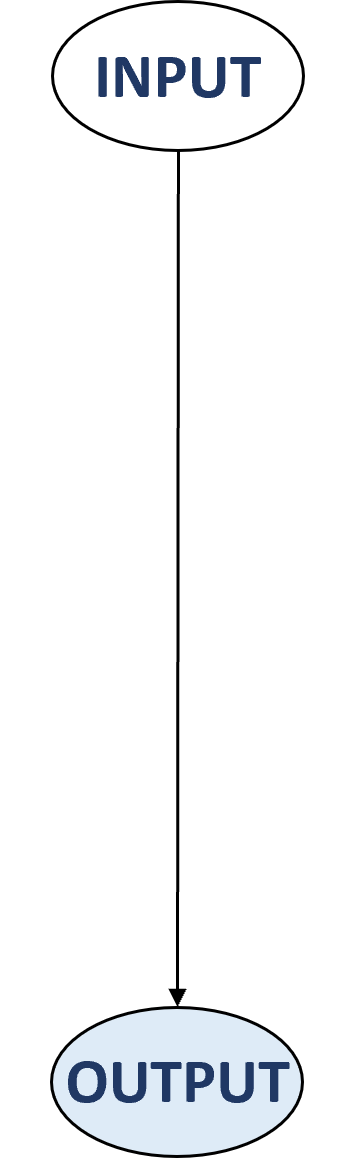}
    }
    \hfill
    \subfigure[CoT]{
        \includegraphics[width=0.20\linewidth]{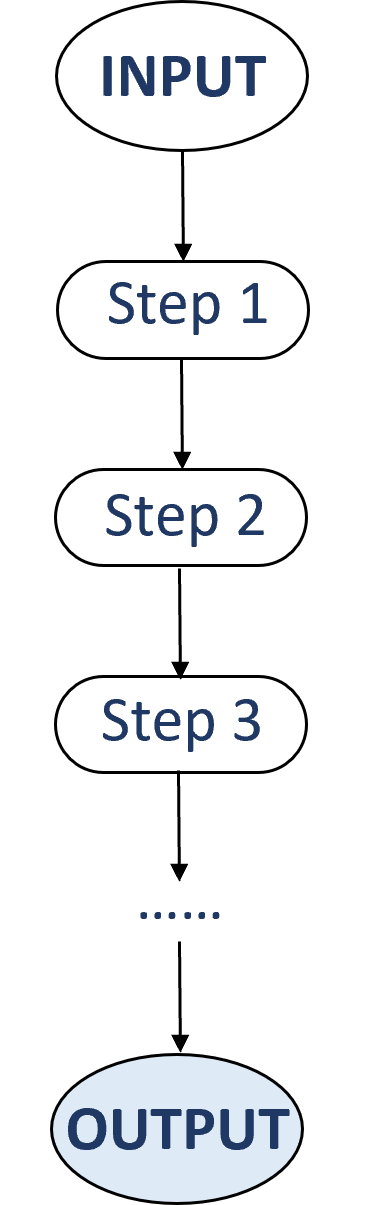}
    }
    \hfill
    \subfigure[ToT]{
        \includegraphics[width=0.47\linewidth]{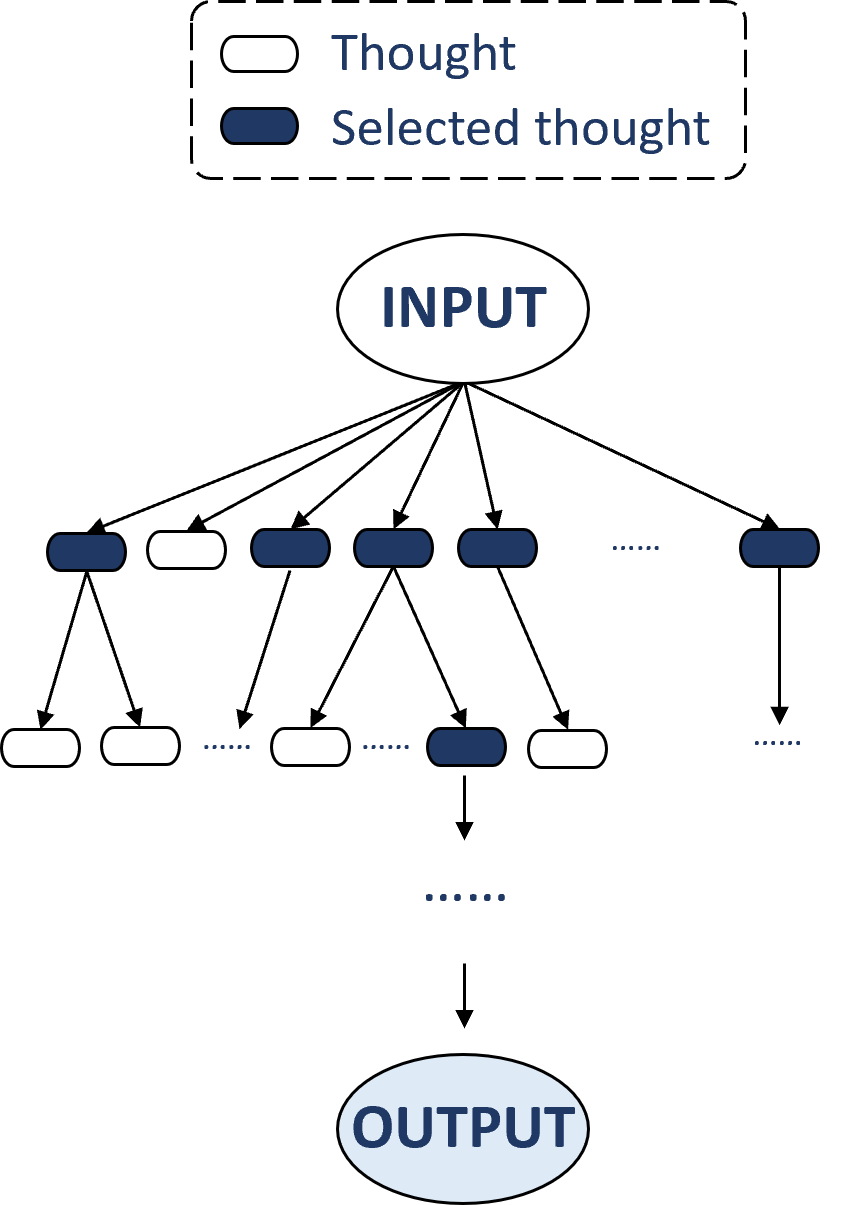}
    }
    \caption{Illustration of IO, CoT and ToT.}
    \label{fig:io cot tot}
\end{figure}

\subsubsection{Baselines}
\paragraph{IO Prompting}
Input-output prompting, or immediate output (also called direct prompting, as shown in Figure \ref{fig:io cot tot} (a)), represents the most direct, fundamental, and commonly used prompting method for guiding LLMs to address problems. In basic IO prompting, the LLM immediately provides a final response upon receiving the initial user prompt, without outputting any intermediate reasoning steps.

\paragraph{CoT}
The CoT method, as illustrated in Figure \ref{fig:io cot tot} (b) and introduced by \citet{COT}, enhances the IO prompting approach by incorporating explicit intermediate reasoning steps beyond the input and output. This method improves the problem-solving capabilities of LLMs, enabling them to address complex problems incrementally.

\subsubsection{ToT Oracle}
In this study, we decompose the ToT framework into two distinct modules — the generator and the discriminator — for separate examination. In order to independently study how each module affects the overall performance of ToT, we utilise an oracle generator and oracle discriminator, respectively, with controllable accuracy to replace the LLM agent of one of the modules. 

The oracle generator, like the LLM agent, produces \(\#generation\) candidate steps, with each step being viable with probability \(p\) (inviable with probability \((1-p)\)). Analogously, the oracle discriminator selects \(\#selection\) steps from the candidate pool using the same approach. 
If no viable step can be generated or selected due to earlier errors, or if the oracle discriminator fails to choose the required step, the task is considered a failure. This process allows us to statistically control the oracle's accuracy at a threshold of \( p \). Additionally, we use a random discriminator that selects \(\#selection\) steps randomly from the candidate pool.\\

Due to the high task complexity of Game of 24, we consistently apply a few-shot prompt in IO, CoT and ToT to help LLMs understand this task. In contrast, for Knights and Knaves, we uniformly employ a zero-shot prompt. For detailed baselines, ToT configurations, and the prompts usage, please refer to Appendix \ref{sec: baseline_tot_setup}.

\section{Experiment and Results}
\subsection{RQ1: Does scaling the size of the \textbf{generator} improve ToT's performance?}\label{sec:RQ1}

\begin{figure}[b!]
    \centering
    \subfigure[Game of 24]{
        \includegraphics[width=\linewidth]{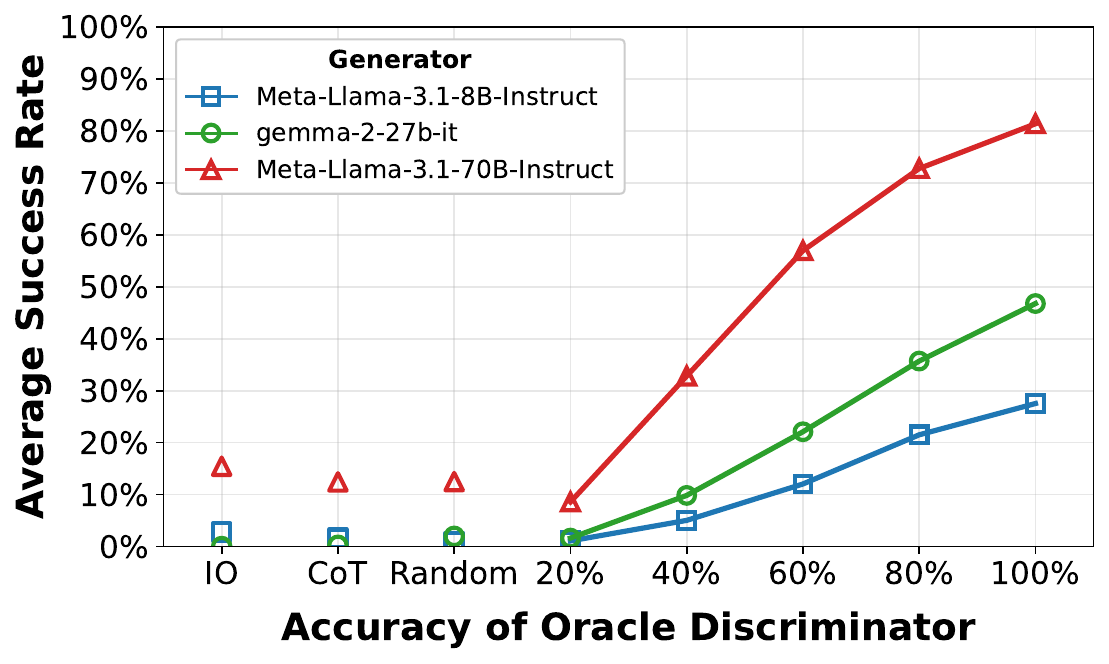}
    }
    % \hfill
    \subfigure[Knights and Knaves]{
        \includegraphics[width=\linewidth]{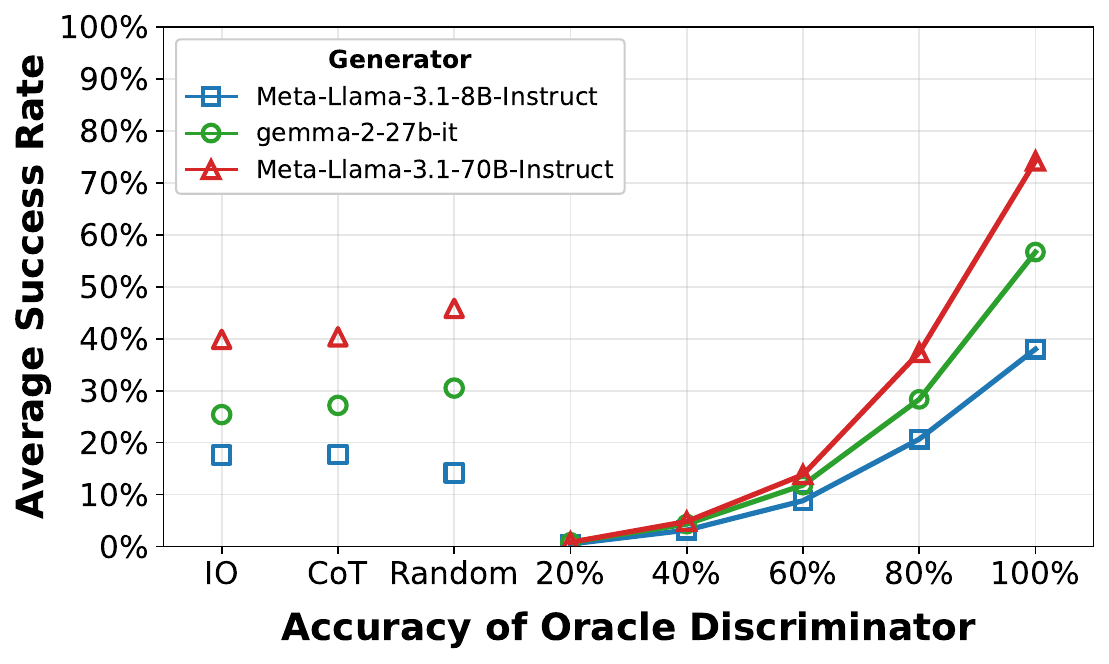}
    }
    \caption{Impact of different models as generators on the overall performance of ToT against oracle discriminators. The lines plot illustrates the average success rate when paired with oracle discriminators. We also plot the performance of IO, CoT, and in combination with a random discriminator.}
    \label{fig:RQ1.1asr}
\end{figure}

\paragraph{Generation Performance Evaluation}
To address RQ1, we employ three different LLMs of varying size as generators and control the accuracy of the oracle discriminator (with accuracy levels of 20\%, 40\%, 60\%, 80\%, 100\%, and a random discriminator) to investigate the impact of the generator while fixing the discriminator at a certain level. At 100\%, we can determine the upper bound of ToT's performance with different generators.

\begin{figure}[h!]
    \centering
    \subfigure[Game of 24]{
        \includegraphics[width=0.95\linewidth]{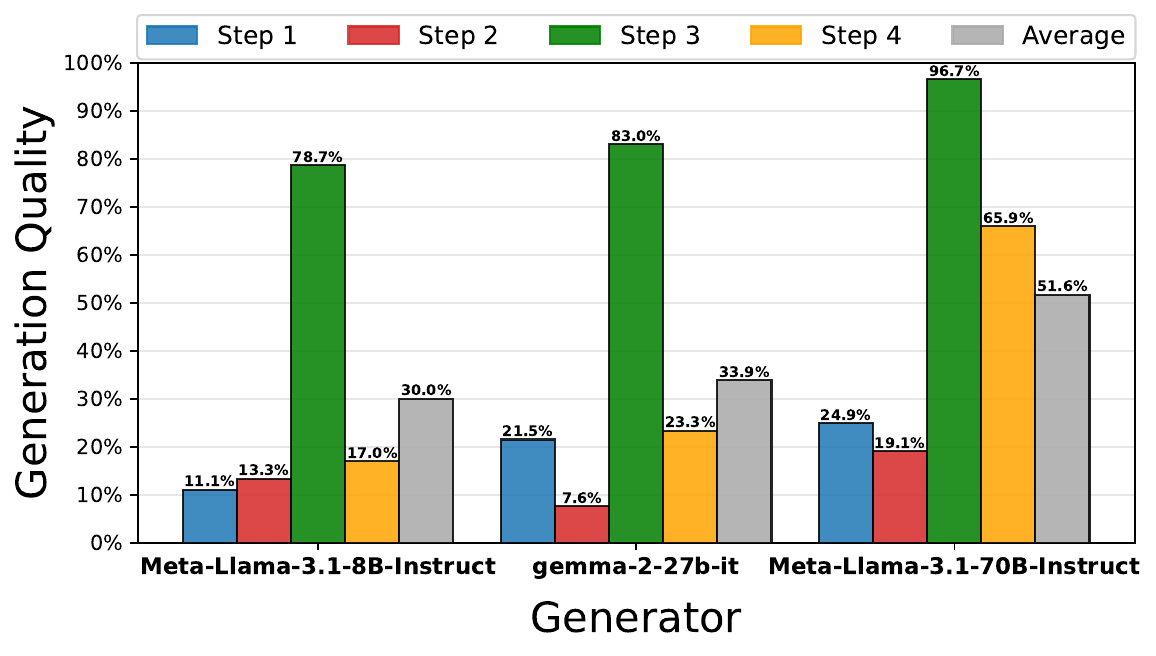}
    }
    % \hfill
    \subfigure[Knights and Knaves]{
        \includegraphics[width=0.95\linewidth]{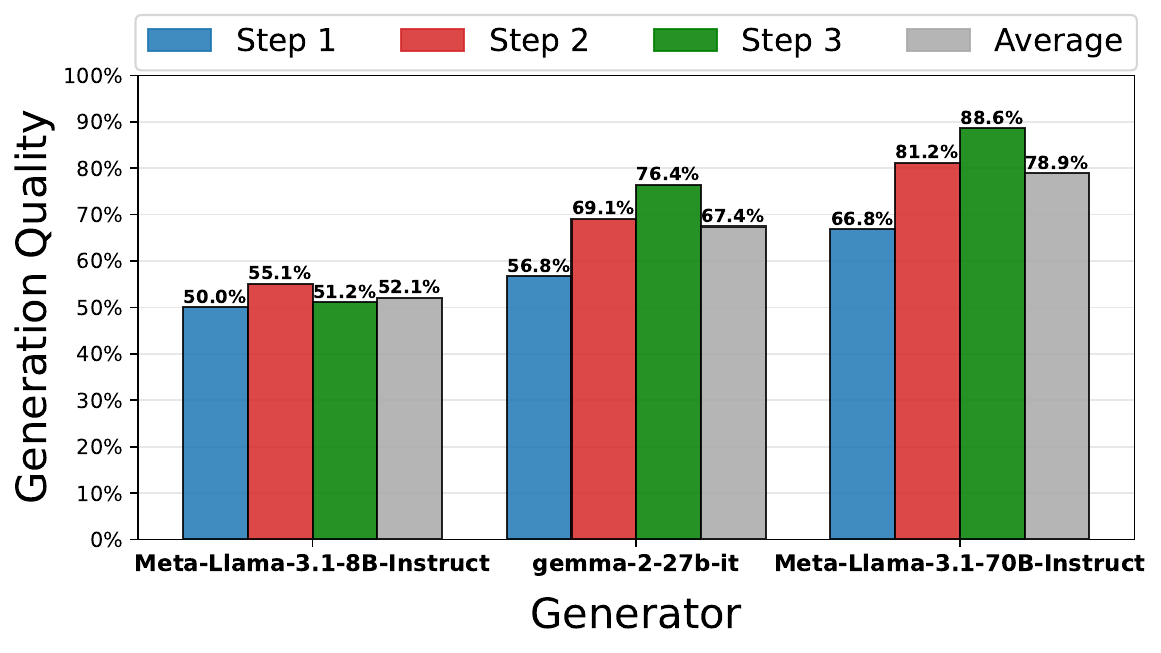}
    }
    \caption{The average generation quality at each step and the overall average generation quality when different models are used as generators in both tasks. The values represent the proportion of unique viable steps among the candidate steps at each step, where higher values indicate better generation quality. We employ the oracle discriminator with 100\% accuracy to eliminate the influence of erroneous preceding steps on subsequent steps, allowing for a clearer analysis of the generator's performance.}
    \label{fig:RQ1.1GQ}
\end{figure}

Figure \ref{fig:RQ1.1asr} displays the performance of LLM generators of varying sizes on Game of 24 and Knights and Knaves, respectively. Besides the performance of the ToT framework, we also plot the performance of the models using IO, CoT, and a random discriminator within the ToT framework as baseline references. The results show that regarding the baselines, CoT is similar to or sometimes even worse than IO on Game of 24. Regarding ToT, for all three LLMs and both datasets, as the performance of the oracle discriminator increases, the overall performance of ToT increases, and substantially outperforms the baseline methods for the Game of 24. For that game, \textbf{the advantages of more powerful generators} (compare Llama-3.1-70B-Inst vs Llama-3.1-8B-Inst) \textbf{become increasingly apparent as the accuracy of the oracle discriminator improves.} A similar trend can be observed in Knights and Knaves, although the difference is less pronounced compared to Game of 24.
Moreover, on Knights and Knaves ToT is often worse than CoT, and only outperforming CoT at high discriminator accuracy levels. 
We hypothesize this is primarily due to the smaller decision complexity of Knights and Knaves in contrast to the larger search space in Game of 24 (see Appendix \ref{sec:g24 braching factor}), which makes it hard to solve in the IO and CoT settings.

\paragraph{Step-wise Generation Quality}
To further investigate the performance differences when using various models as generators, we measure the quality of the intermediate steps generated by each model based on the proportion of unique viable steps among the candidate steps. As shown in Figure \ref{fig:RQ1.1GQ}, \textbf{the performance differences stem from the varying quality of candidate steps generated by different models.} The stronger the model, the higher the proportion of viable steps it generates, statistically increasing the likelihood of selecting viable steps and thus leading to a higher success rate. 

Figure \ref{fig:RQ1.1GQ} (a) illustrates the generation quality of three models acting as generators in the Game of 24. Upon examining the average generation quality at each step individually, we observe that the overall generation quality is low in the first two steps and the final step, remaining below 25\%. However, at the third step, the quality notably increases, with even Llama-3.1-8B-Inst surpassing 78\%. This is because the search space for the first two steps is quite large (48 and 24, respectively), making it difficult to find intermediate steps that result in 24. In contrast, the third step only requires selecting the correct operation sequence, which is relatively straightforward. Furthermore, it is evident that Llama-3.1-70B-Inst demonstrates significantly higher generation quality in the final step compared to smaller models. A detailed analysis of the model responses reveals that smaller models frequently repeat the few-shot examples provided in the prompt when attempting to combine three equations into one. This indicates that merging three equations into one according to the prompt, without being influenced by the prompt examples, presents a challenge for smaller models (see Appendix \ref{sec: baseline_tot_setup} for prompt details).

Figure \ref{fig:RQ1.1GQ} (b) depicts the generation quality of the same three models in Knights and Knaves. It is evident that Llama-3.1-8B-Inst performs similarly to a random generator in Knights and Knaves, with its generation quality only slightly exceeding 50\% at each step. As the model size increases, the overall generation quality also improves progressively.

\subsection{RQ2: Does scaling the size of the \textbf{discriminator} improve the performance of ToT?}\label{sec:RQ2}

\begin{figure}[h!]
    \centering
    \subfigure[Game of 24]{
        \includegraphics[width=0.9\linewidth]{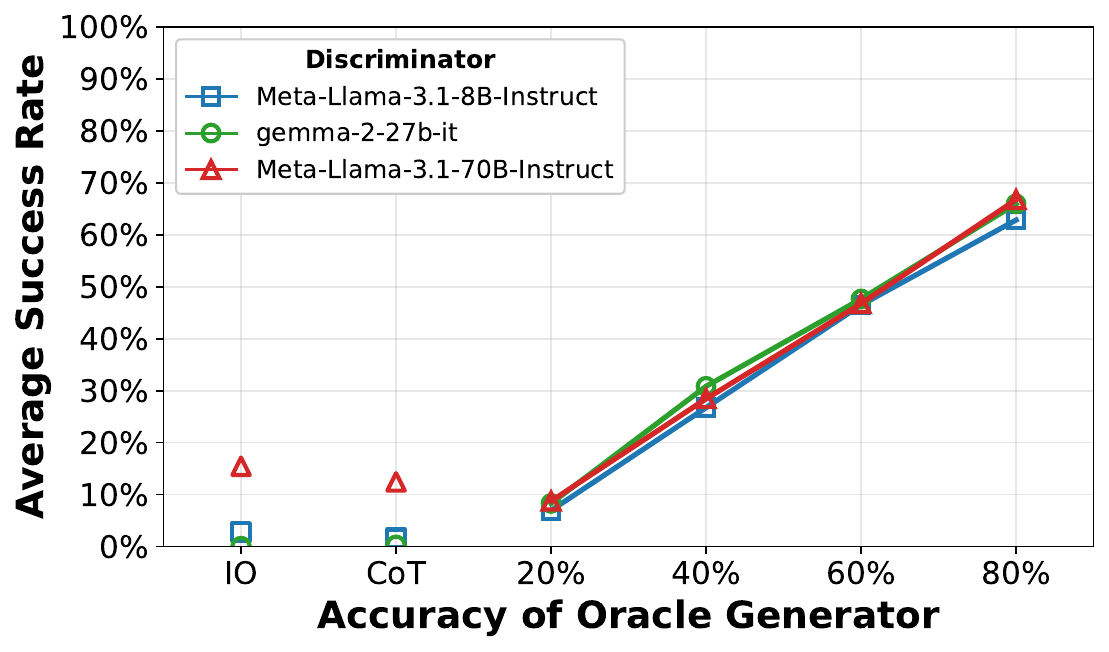}
    }
    % \hfill
    \subfigure[Knights and Knaves]{
        \includegraphics[width=0.9\linewidth]{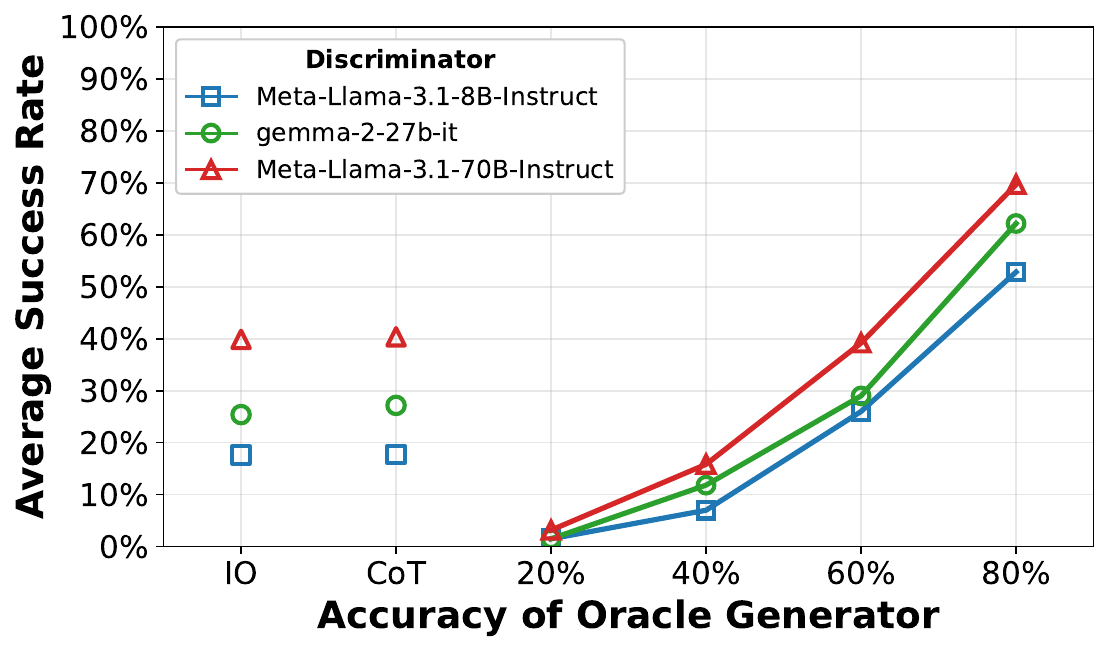}
    }
    \caption{Impact of different models as discriminators on the overall performance of ToT. The lines plot on the right side of the figures illustrate the average success rate when paired with oracle generators. The left side of the figures provide a baseline comparison of the performance of the three models using IO and CoT.}
    \label{fig:RQ2.1}
\end{figure}

\paragraph{Evaluating the Discrimination Performance of LLMs}
To address RQ2, we use three different sizes of LLMs as discriminators and vary the accuracy of the oracle generator (20\%, 40\%, 60\%, 80\%) to control the influence of generators on the experiments. We restrain from using a perfect generator with $100\%$ accuracy as this would make the discriminator redundant.
Unless specified otherwise, in Game of 24, the oracle generator produces 10 candidate steps at each step, while in Knights and Knaves, it generates two, as each character in Knights and Knaves has only two possible identities, whereas the Game of 24 has a much larger game tree.\footnote{For theoretical calculations of the branching factor in Game of 24, refer to the Appendix \ref{sec:g24 braching factor}.} Figure \ref{fig:RQ2.1} displays the performance of different-sized LLMs as discriminators in Game of 24 and Knights and Knaves, respectively. On the left side of the figures, we provide baseline performance using IO and CoT for comparison.

In both tasks, it can be observed that as the performance of the oracle generator improves, the overall effectiveness of ToT also increases. Moreover, similar trends can be observed across the two datasets (with ToT having a lesser impact on Knights and Knaves, further discussed later). Notably, unlike the generator module, the performance differences between models serving as discriminators \textbf{do not significantly widen with the increasing accuracy of the oracle generator}; instead, they remain relatively stable and consistent. In the case of Knights and Knaves, slight differences among the three model sizes emerge when the oracle generator's accuracy exceeds 20\%. However, in the Game of 24, regardless of the oracle generator's accuracy,\textbf{ the three models show almost no difference when used as discriminators}. This suggests that the discriminative ability of Llama-3.1-70B-Inst is not substantially superior to that of Llama-3.1-8B-Inst. Overall, we find that the generator module plays a more significant role in ToT's performance.

\subsection{RQ3: Under which conditions does ToT outperform IO and CoT?}\label{sec:RQ3}

From Figures \ref{fig:RQ1.1asr} and \ref{fig:RQ2.1}, it is evident that ToT does not always outperform baseline methods. Therefore, this section aims to explore the conditions under which LLMs using ToT leads to superior performance compared to baseline approaches.
\paragraph{Task Choice Matters}
As discussed in Sections \ref{sec:RQ1} and \ref{sec:RQ2},
Game of 24 benefits more from using ToT than Knights and Knaves.
As shown in Table \ref{tab:RQ3}, when using a strong generator for Game of 24, ToT performance significantly outperforms the baselines ($33.7\%$ vs. $3.5\%$).
While using a weaker generator, ToT doesn't lead to performance gain, which will be discussed later.
For the baseline methods, they generally fail at such games with a high complexity level.
For instance, in the Game of 24, Llama-3-8B-Inst achieves average success rates below 3\% with IO and CoT setups. Llama-3-70B-Inst achieves a success rate of 3.47\% under IO prompting and 10.44\% under CoT reasoning, as shown in Table \ref{tab:RQ3}. In contrast, in the game of Knights and Knaves, IO and CoT achieve relatively high average success rates. For instance, Llama-3.1-8B-Inst achieves an average baseline success rate of around 17\%, while Llama-3.1-70B-Inst reaches approximately 40\%.
The stronger performance of LLMs using IO and CoT in the Knights and Knaves task can possibly be attributed to its lower decision complexity. 
At each step, the generator only needs to select one out of the available characters (3 maximum) and there are only two possible identities (Truth-teller or Liar).
Whereas in the game of 24, the model needs to select from the available numbers, 4 basic arithmetic operations and the ways to combine them.
The large search space makes it difficult for IO and CoT to fully explore, leading to poor performance.

\begin{table}[h!]
    \centering
    \adjustbox{max width=\linewidth}{
    \begin{tabular}{c|c|c|c}
        \toprule
         Task & Generator & Discriminator & Avg. Success Rate \\
         \midrule
         \multirow{10}{*}{Game of 24} 
         & \multirow{5}{*}{8B} 
         & — (IO) & 2.28\% \\
         && — (CoT) & 1.57\% \\
         && 2B & 0.18\% \\
         && 8B & \textbf{1.29\%} \\
         && 70B & \textbf{1.85\%} \\
        \cline{2-4}
        & \multirow{5}{*}{70B} 
        & — (IO) & 3.47\% \\
        && — (CoT) & 10.44\% \\
        && 2B & 10.76\% \\
        && 8B & \textbf{33.38\%} \\
        && 70B & \textbf{33.76\%} \\
        \midrule
        \multirow{8}{*}{\makecell{Knights\\and\\Knaves}} 
        & \multirow{4}{*}{8B} 
        & — (IO) & 17.63\% \\
        && — (CoT) & 17.70\% \\
        && 8B & \textbf{20.50\%} \\
        && 70B & \textbf{21.33\%} \\
         \cline{2-4}
         & \multirow{4}{*}{70B} 
         & — (IO) & 39.87\% \\
         && — (CoT) & 40.37\% \\
         && 8B & \textbf{52.00\%} \\
         && 70B & \textbf{52.00\%} \\ 
        \bottomrule
    \end{tabular}}
    \caption{Overall performance of ToT when using LLMs as both the generator and discriminator. When fixing the generator, using a larger model as discriminator only gives marginal gain. ToT provides more benefits when using a larger model as a generator. \tablefootnote{In Game of 24, "2B" refers to "gemma-2b-it", "8B" refers to "Llama-3-8B-Inst", and "70B" refers to "Llama-3-70B-Inst". In Knights and Knaves, "8B" refers to "Llama-3.1-8B-Inst" and "70B" refers to "Llama-3.1-70B-Inst"}}
    \label{tab:RQ3}
\end{table}

\paragraph{Strong Generator Matters}
To align the experiments with real-world applications, we conduct experiments using LLMs as both the generator and discriminator modules in ToT and compare them with baseline methods. 
The results, as shown in Table \ref{tab:RQ3}, indicate that \textbf{when the discriminator remains constant, enhancing the generator’s capability can significantly increase the average success rate}. 
In Game of 24, this improvement exceeds 18 times, highlighting that the generator's ability has a greater impact on ToT's performance than the discriminator's. 
 Whereas in Knights and Knaves, enhancing the generator result in a performance increase of over 2.4 times. Additionally, when modifying the discriminator while keeping the same generator, the performance in the Game of 24 improved by no more than 13\%. In Knights and Knaves, there was no observable difference in the discriminatory abilities between the Llama-3-8B-Inst and Llama-3-70B-Inst models. These findings suggest that \textbf{the generator's capacity has a more significant impact on ToT's performance compared to the discriminator. When applying ToT, the advantage of larger models over smaller ones primarily lies in their superior generative capabilities rather than their discriminatory abilities.}

In Game of 24, when the generator is relatively weak (as in the case of Llama-3-8B-Inst), ToT performs significantly worse than IO prompting. However, when the generator is sufficiently strong (as with Llama-3-70B-Inst), ToT can outperform baseline methods, even when gemma-2b-it is used as the discriminator. This finding supports the claim made in Appendix B.2 of \citet{Tree_of_Thoughts} that the bottleneck in Game of 24 lies in step generation. In the Knights and Knaves task, regardless of whether Llama-3.1-8B-Inst or Llama-3.1-70B-Inst is used as the generator, the ToT method consistently outperforms baseline methods. We attribute this difference from the Game of 24 to the lower complexity of Knights and Knaves, which allows Llama-3.1-8B-Inst to act as a generator and still surpass the baseline methods. We believe that if the generator can provide high-quality candidate steps, using a smaller model like Llama-3-8B-Inst as the discriminator is sufficient to significantly enhance the performance of LLMs utilizing ToT, compared to baseline methods.

\section{Related Work}
\paragraph{Scaffolding to Enhance Reasoning of LLMs}
\citet{shwartz-etal-2020-unsupervised} showed that language models can generate chain-of-thought answers with the help of scaffolding by asking clarification questions.
Since the introduction of Scratch Pading \citep{Nye2021ShowYW}, CoT \citep{COT} and zero-shot CoT \citep{kojima2022large}, numerous scaffolding methods have been proposed to further support LLM reasoning. These can be broadly categorized into graph-based approaches — such as SC-CoT \citep{sc-cot}, ToT \citep{Tree_of_Thoughts}, GoT \citep{GOT}, and MindMap \citep{mindmap} — and non-graph-based approaches like AoT \citep{AOT}, IoT \citep{IOT}, XoT \citep{XOT}, PoT \citep{pot} etc. 

\paragraph{Evaluation of Scaffolding Methods}
\citet{gtbench} introduced a novel benchmark aimed at evaluating the strategic reasoning abilities of LLMs in game-theoretic scenarios. 
\citet{topology_of_thoughts_survey} conducted an extensive analysis of various reasoning structures employed by LLMs, exploring the relative strengths and limitations of chain-based, tree-based, and graph-based reasoning strategies. In contrast, our study focuses on an in-depth analysis of the ToT strategy alone, with an emphasis on the capabilities of the generator and discriminator or the complexity of the task.
\citet{discriminator/MCTS} explored the conditions under which tree search methods can enhance the performance of LLMs in planning tasks. Through empirical studies, the authors concluded that the effectiveness of tree search heavily depends on the discriminator's ability to accurately evaluate and guide the search process. Notably, their findings were based on the use of a single generator (i.e., CodeLlama-13B-Inst \citep{codellama-13b}). In contrast, our work employs three different scales of LLMs and an accuracy-controlled oracle generator, providing a more comprehensive investigation into the impact of both the generator and discriminator on the overall performance of the ToT framework.

\section{Conclusion}
This study demonstrates that while ToT offers theoretical advantages, its practical benefits are realized only under specific conditions — namely, when both the generator and discriminator are sufficiently capable. Our findings show that ToT can significantly improve LLM reasoning abilities, but this enhancement depends primarily on the quality of the generator. In tasks with large search spaces, such as Game of 24, stronger generators lead to higher success rates. While the accuracy of the discriminator also contributes to ToT’s performance, stronger LLMs do not provide superior discrimination performance. Therefore, we recommend using a state-of-the-art large model as the generator alongside a smaller model as the discriminator to harness the advantages of ToT while significantly reducing computational costs, without the need for extreme performance optimization.
Furthermore, the benefits of ToT are more pronounced in highly complex tasks where methods like IO and CoT show poor performance, emphasizing the value of using ToT in such challenging scenarios.

\section{Limitations}
This study has several limitations. First, we only experimented with the Knights and Knaves subset containing three characters, which led to overly optimistic performance for the baseline methods, partially masking the advantages of ToT. Second, the parameter settings for ToT are another important factor influencing its effectiveness. Future work should expand the range and complexity of tasks and consider the impact of ToT's parameter settings to gain a more comprehensive understanding of its effectiveness.

\section*{Acknowledgments}
We thank the members of MaiNLP labs for their constructive feedback.
We especially appreciate the suggestions of Siyao Peng, Diego Frassinelli, Shijia Zhou, Bolei Ma and Rob van der Goot.
This research is supported by ERC Consolidator Grant DIALECT 101043235.

\bibliography{custom}

\appendix

\section{Experimental Setup}
\subsection{Baselines \& ToT Setup}\label{sec: baseline_tot_setup} % TODO 
\subsubsection{Game of 24}\label{sec: method_g24_setup}
Our implementation is largely based on \citet{Tree_of_Thoughts}'s work. However, during our pilot experiment, we found that the same setup did not allow the selected open-source model to understand the task well. Therefore, we made some adjustments based on \cite{Tree_of_Thoughts}'s work.

In terms of IO prompting, we follow \cite{Tree_of_Thoughts}'s approach of using five in-context examples to support the IO prompt, but we adjust the order of the examples. The IO prompt is as follows:

\begin{tcolorbox}[colback=white!0!white,colframe=black!0!black,title=IO Prompt,fontupper=\footnotesize, enhanced, breakable]
Use numbers and basic arithmetic operations (+ - * /) to obtain 24.\\
Input: 2 9 10 12

Answer: 2 * 12 * (10 - 9) = 24

Input: 4 9 10 13

Answer: (13 - 9) * (10 - 4) = 24

Input: 1 4 8 8

Answer: (8 / 4 + 1) * 8 = 24

Input: 5 5 5 9

Answer: 5 + 5 + 5 + 9 = 24

Input: 4 4 6 8

Answer: (4 + 8) * (6 - 4) = 24

Input: <input>
\end{tcolorbox}

As for CoT, we use the same prompt as \cite{Tree_of_Thoughts}, add three intermediate equations in each input-output pair, and present five examples to LLMs. The CoT prompt is as follows:

\begin{tcolorbox}[colback=white!0!white,colframe=black!0!black,title=CoT Prompt,fontupper=\footnotesize, enhanced, breakable]
Use numbers and basic arithmetic operations (+ - * /) to obtain 24. Each step, you are only allowed to choose two of the remaining numbers to obtain a new number.\\
Input: 4 4 6 8\\
Steps:\\
4 + 8 = 12 (left: 4 6 12)\\
6 - 4 = 2 (left: 2 12)\\
2 * 12 = 24 (left: 24)\\
Answer: (6 - 4) * (4 + 8) = 24\\
Input: 2 9 10 12\\
Steps:\\
12 * 2 = 24 (left: 9 10 24)\\
10 - 9 = 1 (left: 1 24)\\
24 * 1 = 24 (left: 24)\\
Answer: (12 * 2) * (10 - 9) = 24\\
Input: 4 9 10 13\\
Steps:\\
13 - 10 = 3 (left: 3 4 9)\\
9 - 3 = 6 (left: 4 6)\\
4 * 6 = 24 (left: 24)\\
Answer: 4 * (9 - (13 - 10)) = 24\\
Input: 1 4 8 8\\
Steps:\\
8 / 4 = 2 (left: 1 2 8)\\
1 + 2 = 3 (left: 3 8)\\
3 * 8 = 24 (left: 24)\\
Answer: (1 + 8 / 4) * 8 = 24\\
Input: 5 5 5 9\\
Steps:\\
5 + 5 = 10 (left: 5 9 10)\\
10 + 5 = 15 (left: 9 15)\\
15 + 9 = 24 (left: 24)\\
Answer: ((5 + 5) + 5) + 9 = 24\\
Input: <input>
\end{tcolorbox}

Regarding the ToT approach, we emulate the steps of the CoT process. In each step, the generator produces multiple candidate steps. The discriminator then \underline{evaluates} each candidate step \underline{three times} and \underline{selects} the \underline{top five} to proceed to the next step. Consequently, the LLM generates up to five answers for each task. For fairness, we also instruct the LLM to generate five answers when using the baseline methods. To mitigate the inflated success rate from multiple attempts, we use average success rate (see Section \ref{sec:metric}) alongside overall success rate.

In pilot experiments, we observe that some small open-source models struggled to generate valid intermediate steps, resulting in inefficient ToT performance and poor experimental outcomes. To address this, we introduce a filter in the generation step to discard obviously invalid intermediate steps (e.g., steps with mismatched remaining numbers), before passing the filtered candidates to the discriminator for evaluation. This not only enhancs ToT efficiency but also improves task success rates to a certain extent.

For the generator, we use a three-shot generation prompt to produce equations for the first three steps, and a five-shot merge prompt to consolidate the intermediate steps into a single equation in the final step. 
For the discriminator, we similarly employ a three-shot value prompt to guide LLMs in assessing the potential of completing the game in the first three steps, and another zero-shot value prompt to evaluate the correctness of the final step. The four prompts are as follows.

\begin{tcolorbox}[colback=white!0!white,colframe=black!0!black,title=ToT: Generation Prompt,fontupper=\footnotesize, enhanced, breakable]
Input: 2 8 8 14\\
Possible next steps:\\
2 + 8 = 10 (left: 8 10 14)\\
8 / 2 = 4 (left: 4 8 14)\\
14 + 2 = 16 (left: 8 8 16)\\
2 * 8 = 16 (left: 8 14 16)\\
8 - 2 = 6 (left: 6 8 14)\\
14 - 8 = 6 (left: 2 6 8)\\
14 / 2 = 7 (left: 7 8 8)\\
14 - 2 = 12 (left: 8 8 12)\\
Input: 4 4 10\\
Possible next steps:\\
4 + 4 = 8 (left: 8 10)\\
4 * 10 = 40 (left: 4 40)\\
10 - 4 = 6 (left: 4 6)\\
4 / 4 = 1 (left: 1 10)\\
4 - 4 = 0 (left: 0 10)\\
4 + 10 = 14 (left: 4 14)\\
4 * 4 = 16 (left: 10 16)\\
Input: 10 14\\
Possible next steps:\\
10 + 14 = 24 (left: 24)\\
14 - 10 = 4 (left: 4)\\
10 * 14 = 140 (left: 140)\\
14 / 10 = 1.4 (left: 1.4)\\
Generate possible next steps for the following inputs, following the example above. Note that the number of leftover digits should be one less than the number of input digits.\\
Input: <input>\\
Possible next steps:
\end{tcolorbox}

\begin{tcolorbox}[colback=white!0!white, colframe=black!0!black, title=ToT: Merge Prompt, fontupper=\footnotesize, enhanced, breakable]
Given three calculation steps. Follow the examples and combine the three calculation steps into one equation, but do not simplify. Your output should be in this format "Answer: \{combined one equation\}"\\
Examples:\\
Steps:\\
8 / 4 = 2 (left: 1 2 8)\\
1 + 2 = 3 (left: 3 8)\\
3 * 8 = 24 (left: 24)\\
Answer: (1 + 8 / 4) * 8 = 24\\
Steps:\\
12 * 2 = 24 (left: 9 10 24)\\
10 - 9 = 1 (left: 1 24)\\
24 * 1 = 24 (left: 24)\\
Answer: (12 * 2) * (10 - 9) = 24\\
Steps:\\
4 + 8 = 12 (left: 4 6 12)\\
6 - 4 = 2 (left: 2 12)\\
2 * 12 = 24 (left: 24)\\
Answer: (6 - 4) * (4 + 8) = 24\\
Steps:\\
5 + 5 = 10 (left: 5 9 10)\\
10 + 5 = 15 (left: 9 15)\\
15 + 9 = 24 (left: 24)\\
Answer: ((5 + 5) + 5) + 9 = 24\\
Steps:\\
13 - 10 = 3 (left: 3 4 9)\\
9 - 3 = 6 (left: 4 6)\\
4 * 6 = 24 (left: 24)\\
Answer: 4 * (9 - (13 - 10)) = 24\\
It's your turn:\\
Steps: <steps>
\end{tcolorbox}

\begin{tcolorbox}[colback=white!0!white,colframe=black!0!black,title=ToT: Value Prompt,fontupper=\footnotesize, enhanced, breakable]
Evaluate if given numbers can reach 24. Conclude in the last line "confident", "likely" or "impossible".\\
Example 1:\\
4 4 10\\
4 + 4 + 10 = 8 + 10 = 18\\
4 * 10 - 4 = 40 - 4 = 36\\
(10 - 4) * 4 = 6 * 4 = 24\\
confident\\
Example 2:\\
11 12\\
11 + 12 = 23\\
12 - 11 = 1\\
11 * 12 = 132\\
11 / 12 = 0.91\\
impossible\\
Example 3:\\
5 7 8\\
5 + 7 + 8 = 12 + 8 = 20\\
(8 - 5) * 7 = 3 * 7 = 21\\
I cannot obtain 24 now, but numbers are within a reasonable range\\
likely\\
Now you should evaluate:\\
<input>
\end{tcolorbox}

\begin{tcolorbox}[colback=white!0!white,colframe=black!0!black,title=ToT: Value Prompt for last step, fontupper=\footnotesize, enhanced, breakable]
Given an input and an answer, evaluate if the answer is correct, i.e. it uses each input exactly once and no other numbers, calculation is correct and reaches 24.Give your judgement in the last line: "confident" or "impossible".\\
Input: <input>\\
Answer: <answer>
\end{tcolorbox}

\subsubsection{Knights and Knaves}
This study adopts \citet{mondorf2024liarliarlogicalmire}'s zero-shot IO- and CoT prompts as baselines. The IO prompt consists of a system message and an instruction, while the CoT prompt extends the IO prompt by adding "Let's think step by step."
\begin{tcolorbox}[colback=white!0!white,colframe=black!0!black,title=System Message, fontupper=\footnotesize, enhanced, breakable]
Your task is to solve a logical reasoning problem. You are given set of statements from which you must logically deduce the identity of a set of characters.\\

You must infer the identity of each character. First, explain your reasoning. At the end of your answer, you must clearly state the identity of each character by following the format:\\

CONCLUSION:\\
A: ...\\
B: ...\\
C: ...\\
...
\end{tcolorbox}
\begin{tcolorbox}[colback=white!0!white,colframe=black!0!black,title=Instruction, fontupper=\footnotesize, enhanced, breakable]
\#\#\# Instruction \#\#\#\\
Assume that there exist only two types of people: truth-tellers and liars. truth-tellers always tell the truth, while liars always lie.\\
You are given the statements from <num-characters> characters. Based on their statements, infer who is a truth-teller and who is a liar.\\

Based on the following statements, infer who is a truth-teller and who is a liar:\\
<statements>\\

First, explain your reasoning. End your answer by clearly stating the identity of each character in the following format:\\

A: truth-teller/liar\\
B: truth-teller/liar\\
C: truth-teller/liar\\
...
\end{tcolorbox}

In the ToT setup, we have the LLM agent analyze each character's identity step by step, where reasoning at each subsequent step is based on inferences from prior steps. In this way, the Knights and Knaves problem can be viewed as a full binary tree with a depth equal to the number of characters. The two nodes at each level represent the two possible identities of a character (truth-teller/liar), and the unique solution corresponds to one root-to-leaf path. Since each character has only two possible identities, the default \underline{number of generated steps is set to 2}. The discriminator uses a \underline{three-round voting} mechanism to select the best step from the two candidates. In practice, LLMs allows the generation of two steps with the same conclusion. The prompt for Knights and Knaves is as follows, we make it zero-shot to be consistent with baseline setup:

\begin{tcolorbox}[colback=white!0!white,colframe=black!0!black,title=ToT: Generation Prompt, fontupper=\footnotesize, enhanced, breakable]
\#\#\# Instruction \#\#\#\\
Assume that there exist only two types of people: truth-tellers and liars. truth-tellers always tell the truth, while liars always lie.\\
You are given the statements from <num-characters> characters. Based on their statements, and some known identities, infer who is a truth-teller and who is a liar.\\

Statements:\\
<statements>\\

Known identities:\\
<known\_identities>\\

Now, infer the identity of <character> and explain your reasoning. End your answer by clearly stating the identity of <character> in the following format:\\

<character>: truth-teller/liar
\end{tcolorbox}
\begin{tcolorbox}[colback=white!0!white,colframe=black!0!black,title=ToT: Vote Prompt, fontupper=\footnotesize, enhanced, breakable]
Given an instruction, several statements, known identities and several choices, decide which choice is most promising.\\
Analyze each choice in detail, then conclude in the last line "The best choice is \{s\}", where s the integer id of the choice.\\

\#\#\# Instruction \#\#\#\\
Assume that there exist only two types of people: truth-tellers and liars. truth-tellers always tell the truth, while liars always lie.\\
You are given the statements from <num-characters> characters. Based on their statements, and some known identities, infer who is a truth-teller and who is a liar.\\

Statements:\\
<statements>\\

Known identities:\\
<known\_identities>\\

Choice 1: <first candidate step>\\
Choice 2: <second candidate step>
\end{tcolorbox}

Unless otherwise specified, we use the subset of the Knights and Knaves task with 3 characters.

\subsection{Hardware}
Our experiments are conducted on NVIDIA-A100 GPUs, each with either 40 GB or 80 GB of memory.

\section{Theoretical Calculation of the Branching Factor in the Game of 24 Game Tree}\label{sec:g24 braching factor}
\begin{enumerate}
    \item In the first step, players can select two numbers from four, and choose one from the four basic arithmetic operators to perform calculations. Since subtraction and division do not obey the commutative law, if we consider equivalent operations that obey the commutative law as different choices, then the exploration space for the first step is $4 \times 3 \times \binom{4}{1} = 48$.
    \item In the second step, players can select two numbers from the remaining three, and choose one from the four basic arithmetic operators to perform calculations. Therefore, the exploration space for the second step is $3 \times 2 \times \binom{4}{1} = 24$.
    \item In the third step, only two numbers remain. Players need to choose one from the four basic arithmetic operators to perform calculations. Therefore, the exploration space for the third step is $2 \times 1 \times \binom{4}{1} = 8$.
    \item In the fourth step, players only need to integrate the previous three calculation steps into one formula. We assume that the player combines the three expressions based on the sequence of the first three intermediate steps, without considering the commutative property or equivalent operations using parentheses. Therefore, the exploration space for the fourth step is 1.
\end{enumerate}
In summary, the branching factor for Game of 24 is \{1, 8, 24, 48\}.

It is important to note that our calculations are based on the assumption that the four initial numbers are completely distinct (e.g., 1, 2, 3, 4). If there are repetitions among the four numbers (e.g., 1, 8, 8, 8, or 6, 6, 6, 6), the actual branching factor would be smaller. The specific branching factor depends on the nature of the repetitions. Given that the repetition of remaining numbers in the intermediate steps largely depends on the calculations considered by the player, we simplify our computation by assuming all numbers are distinct. Consequently, the branching factor we obtain represents the upper limit for all possible scenarios.

\section{Algorithm for generating complete answers to Game of 24}\label{sec:algo}
\begin{algorithm}
\caption{Solve 24 Game}
\begin{algorithmic}[1]
\Function{solve\_24\_game}{numbers, target = 24}
    \If{numbers is empty}
        \State \Return empty list
    \EndIf
    \State Initialize solutions as empty list
    \State Set $\epsilon$ to small tolerance value for floating point comparison

    \Function{dfs}{nums, steps}
        \If{length of nums = 1}
            \If{nums[0] is approximately target within tolerance $\epsilon$}
                \State Add current steps to solutions
            \EndIf
            \State \Return
        \EndIf

        \For{each pair $(a, b)$ from nums}
            \If{$a = b$}
                \State \textbf{continue}
            \EndIf
            \State Create next\_nums by removing $a$ and $b$ from nums
            
            \State Define operations list as:
            \State $a + b, a - b, b - a, a * b$
            \If{$b \neq 0$}
                \State Add $a / b$ to operations
            \EndIf
            \If{$a \neq 0$}
                \State Add $b / a$ to operations
            \EndIf

            \For{each operation result, expression in operations}
                \State Append expression to steps
                \State Call dfs(next\_nums + result, updated steps)
            \EndFor
        \EndFor
    \EndFunction

    \State Call \texttt{dfs}(numbers, empty steps)
    \State \Return solutions
\EndFunction
\end{algorithmic}
\end{algorithm}

\end{document}